\documentclass{article}

\PassOptionsToPackage{numbers, compress}{natbib}

\usepackage[preprint]{neurips_2025}

\usepackage[utf8]{inputenc} 
\usepackage[T1]{fontenc}    
\usepackage{hyperref}       
\usepackage{url}            
\usepackage{booktabs}       
\usepackage{amsfonts}       
\usepackage{nicefrac}       
\usepackage{microtype}      
\usepackage{xcolor}         

\usepackage{tikz}
\usepackage{amsthm}

\usepackage{algorithm}
\usepackage[noend]{algorithmic}
\usepackage{amsmath}        
\usepackage{multirow} 

\newcommand{\g}{\ensuremath{\mathcal{G}}}
\newcommand{\V}{\ensuremath{\mathbf{V}}}
\newcommand{\X}{\ensuremath{\mathbf{X}}}

\newcommand{\U}{\ensuremath{\mathbf{U}}}
\newcommand{\f}{\ensuremath{\mathbf{f}}}

\newtheorem{definition}{Definition}
\newtheorem{proposition}{Proposition}
\newtheorem{lemma}{Lemma}

\title{Improving Generalization Ability of \\Robotic Imitation Learning by Resolving \\Causal Confusion in Observations}

\author{
  Yifei Chen\thanks{\texttt{All authors are with DATA61 of CSIRO, Australia. \\Email addresses: \{Yifei.Chen, Yuzhe.Zhang, Giovanni.Durso, Nicholas.Lawrance, Brendan.Tidd\}@data61.csiro.au}} \quad Yuzhe Zhang \quad Giovanni Durso \quad Nicholas Lawrance \quad Brendan Tidd \\
}

\begin{document}
\maketitle

\begin{abstract}
Recent developments in imitation learning have considerably advanced robotic manipulation.
However, current techniques in imitation learning can suffer from poor generalization, limiting performance even under relatively minor domain shifts.
In this work, we aim to enhance the generalization capabilities of complex imitation learning algorithms to handle unpredictable changes from the training environments to deployment environments.
To avoid confusion caused by observations that are not relevant to the target task, we propose to explicitly learn the causal relationship between observation components and expert actions, employing a framework similar to \cite{de2019causal}, where a causal structural function is learned by intervention on the imitation learning policy.
Disentangling the feature representation from image input as in \cite{de2019causal} is hard to satisfy in complex imitation learning process in robotic manipulation, we theoretically clarify that this requirement is not necessary in causal relationship learning.
Therefore, we propose a simple causal structure learning framework that can be easily embedded in recent imitation learning architectures, such as the Action Chunking Transformer \cite{zhao2023learningACT}.
We demonstrate our approach using a simulation of the ALOHA~\cite{zhao2023learningACT} bimanual robot arms in Mujoco, and show that the method can considerably mitigate the generalization problem of existing complex imitation learning algorithms.

\end{abstract}


\section{Introduction}\label{sec:intro}
Imitation learning (IL) is a powerful framework for robotics, particularly manipulation. This technique allows agents to learn complex skills directly from expert demonstrations without explicitly defining the task or reward~\cite{osaAlgorithmicPerspective2018}.
Transformer-based methods, such as the Action Chunking Transformer (ACT)~\cite{zhao2023learningACT} have shown promise due to their ability to handle high-dimensional problems and model temporal dependencies, enabling the completion of a diverse range of tasks. 
Despite this advancement, their lack of robustness under distribution shift of task or environment impacts their real-world applicability.

One key source of this weak generalization is \emph{causal confusion}, where a model learns spurious correlations between what caused an effect and the result of the effect on the environment~\cite{Pearl2000causal, peters2017elements}.
This manifests itself as a mapping between task-relevant \emph{causal} features and irrelevant features, such as focusing on the color of the background or other unimportant objects in the scene. An example is a model that learns to classify hair color and only focuses on gender as a predictive feature~\cite{izmailov2022feature}.
Unfortunately, if the model is tested in a new environment without these irrelevant features and repeats the same task the performance can degrade sharply due to this misaligned learning.
Prior work by \cite{de2019causal} attempted to solve this problem using causal structure learning, but requires a strong assumption that all observations in the environment are disentangled, which limits its use in robotics.

In this work, we propose \textbf{Causal-ACT}, an extension to IL models that integrates causal structure learning into a transformer-based policy model.
Our method avoids the reliance on disentangled representations and learns a causal structure model from an off-the-shelf convolutional encoder, such as ResNet-18~\cite{targ2016resnet}. 
Our method jointly optimizes over a space of candidate graphs and intervenes to find the appropriate structure that learns the task-relevant features and avoids causal confusion.
This allows the policy to focus on what matters and improves its robustness to domain shifts at test time.

We demonstrate our algorithm's effectiveness in simulated experiments using the ALOHA bimanual robot to complete a manipulation task. For these experiments we train the policies with task-irrelevant features in the environment then test in an out-of-distribution environment with the irrelevant features removed. Our results show that our Causal-ACT method substantially outperforms the traditional ACT method under distribution shift conditions. Additionally, we demonstrate that the approach has similar performance to ACT with domain randomization without the need for additional expert demonstrations or computational requirements.


The contributions of this work are:
We identify and address the causal confusion problem in Imitation Learning. We provide a theoretical justification for relaxing the disentangled requirement and prove under which graphical conditions you can learn causal structural functions for imitation learning. We develop Causal-ACT, an example of using causal learning in imitation learning without needing the restriction of disentangled embeddings. We empirically prove Causal-ACTS performance through simulation and demonstrate its ability to improve generalization\footnote{We defer the theoretical proof, and additional experiment details to the Appendix. Our code can be found in this anonymous repo: \href{https://anonymous.4open.science/r/Causal_ACT_code-E7BA/README.md}{https://anonymous.4open.science/r/Causal\_ACT\_code-E7BA/README.md}.}.

\section{Related work}
\paragraph{Imitation learning for manipulation tasks.}
Imitation learning is often used for tasks in which defining the reward function explicitly is difficult to define or may even have subjective qualities~\cite{osaAlgorithmicPerspective2018, hoGenerativeAdversarial2016}. If there are expert demonstrations available, they can be used to guide the learning of policies to complete the task directly. 
Transformer-based models~\cite{shafiullahBehaviorTransformers2022, brohanRT2VisionLanguageAction2023, shridharPerceiverActorMultiTask2023} have been shown to perform well in IL tasks; performance in robotic manipulation tasks can be further improved through the use of `action chunking tasks'~\cite {zhao2023learningACT, black2410pi0}.
Although these methods can deal with some environmental mismatch between training and deployment, they are still vulnerable to large enough domain discrepancies~\cite{zareSurveyImitation2024}.
\paragraph{Methods for Improving Generalization in Robotics.}
To improve generalization, one dominant strategy is large-scale data collection from diverse real-world environments~\cite{zhao2024aloha, o2024open}, but this is often impractical due to high costs in time, labor, and computation.
An alternative approach to large-scale real-world data collection is using simulated environments to augment or replace real-world data. domain randomization (DR)~\cite{tobinDomainRandomization2017, makoviychukIsaacGym2021, anderson2021sim} and generative simulation methods~\cite{jamesSimToRealSimToSim2019} attempt to expose models to various conditions. In particular, DR varies parameters such as lightning, object positions, textures, and friction.
However, the quality of the learned models strongly depends on the chosen tuning parameters, quality and quantity of the simulated data~\cite{prakashStructuredDomain2019}, costs additional computational resources~\cite{openaiSolvingRubiks2019}, and can result in overly conservative policies~\cite{zhao2020sim}. This results in methods that are fragile in different ways and are computationally expensive to train due to their sample inefficiency.

\paragraph{Causal Methods for Robust Learning.}
Causal methods have been proposed as a technique to improve the robustness and generalization ability of learning-based methods by modeling the causal relationships between observations and actions~\cite{Pearl2000causal, peters2017elements}. This helps distinguish task-relevant features from spurious correlations, reducing overfitting to irrelevant training patterns and increasing sample efficiency~\cite{yaoSurveyCausal2021}. As a result, models become more resilient to domain shifts. These approaches have shown promise in visual prediction~\cite{zengTransporterNetworks2021}, reinforcement learning~\cite{zhang2020invariant}, and imitation learning~\cite{de2019causal}, offering a structured alternative to data-heavy strategies.\\
The closest related work to our approach is the causal confusion framework proposed by de Haan et al.~\cite{de2019causal}, which aims to identify causally relevant features from disentangled latent representations learned by a Variational Autoencoder ($\beta$-VAE). However, this introduces inductive biases~\cite{locatello2019challenging} and is non-trivial to learn for complex robotic manipulation settings with high-dimensional observations and large, coupled action spaces. In contrast, our work removes the need for disentanglement, which enables improved generalization to out-of-distribution scenarios without relying on extensive data collection.

\section{Insufficient generalizability of imitation learning policies}\label{sec:confusion}
\paragraph{Imitation learning} In imitation learning, a robot tries to learn a {\em policy}, i.e., a function $\pi: \mathcal{X}\rightarrow \mathcal{A}$ taking an {\em observation} $\X^t$ from the observation space $\mathcal{X}$ and outputting an {\em action} $A^t$ from the corresponding space $\mathcal{A}$ for each time step $t$, so that executing the policy generates a temporal sequence of observations and actions that is as close to a set of given {\em demonstrations} as possible.
We assume that an observation $\X^t=(X^t_1, \cdots, X^t_n)$ is an $n$-dimensional vector, and it can be camera images or their embedded representation (the output of an image encoding network such as a ResNet-50 penultimate layer).

Note that in the above definition, we assume that at each time step, the policy $\pi$ predicts a single-step action, i.e., $A^t=\pi(\X^t)$.
However, $\pi$'s output can be a sequence of consecutive actions, e.g., in the ACT algorithm \cite{zhao2023learningACT}.

\paragraph{A simple example of performance drop facing domain discrepancy}

\begin{figure}[t]
     \centering
     \includegraphics[width=0.8\textwidth, trim=1cm 2cm 1cm 2cm, clip]{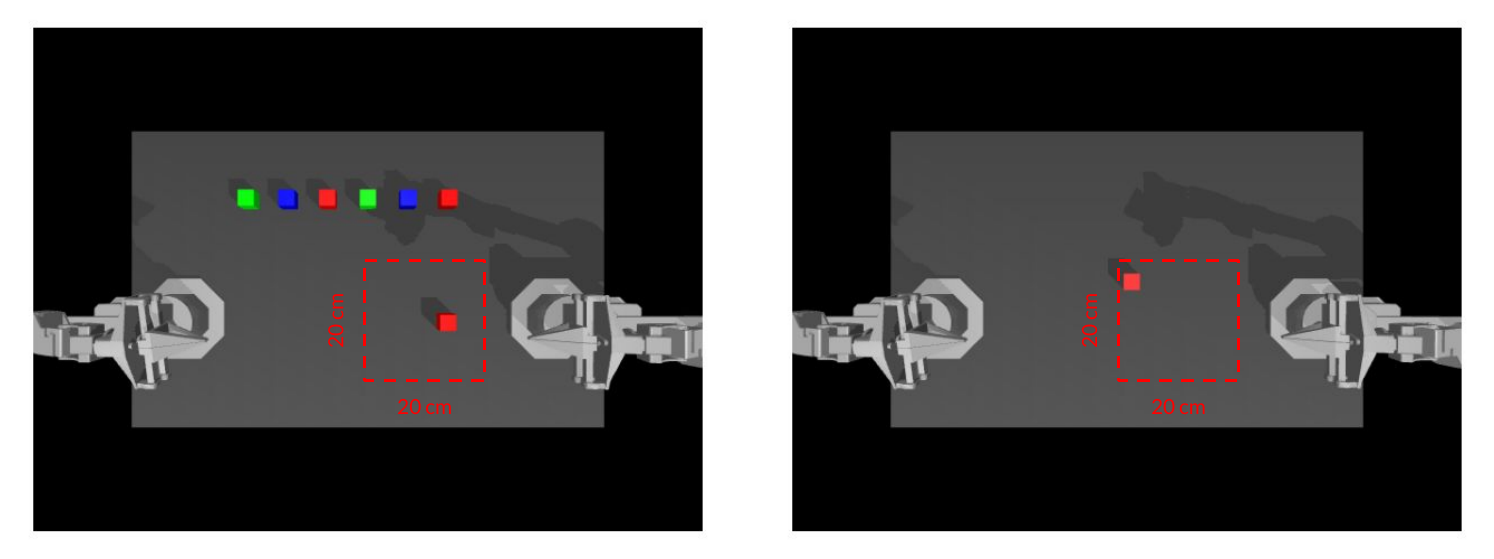}
     \caption{A visualization of the training (left) and the testing (right) environment. Note that the red squares and texts on both figures only indicate the extent of the sampling range for the goal cube initial position and are not included in observations.}
     \label{fig:env}
\end{figure}


To concretely demonstrate our motivation, we take ACT \cite{zhao2023learningACT} as an example to illustrate its vulnerability facing a domain discrepancy from the training environment to the test environment.
Consider the same task described in our experiment (Section \ref{sec:experiment}), that is, two end-effectors of an ALOHA robot are trained to grasp the target cube (the red one initially located in the red square) with one arm and transfer it to the other. However, from the training environment to the test environment (i.e., from left to right of Figure \ref{fig:env}), we remove the six distracting cubes to model an unpredictable change in deployment environment.
We observe a drastic performance drop when the robot faces such a mild environment change.


Without proper handling, the robot learns to make decisions based not only on the target, but also on the surroundings.
However, the surrounding environment should not be a factor in deciding the next action; instead, the shape and location of the target cube are the real factors.
Relying on such task-irrelevant information can obstruct the policy from properly responding to changes in the surroundings.
Therefore, it is crucial for the robot to learn which observations decide the next action from the demonstration sequences such that it can learn generalizable skill knowledge instead of a mapping that overfits to the training environment.


\section{Resolving causal confusion enhances generalizability}

\subsection{Modeling the imitation policy as a causal model}\label{sec:causal_model}
\begin{figure}[t]
\centering
\resizebox{0.6\textwidth}{!}{ 
\begin{tikzpicture}
\node (d1) at (-6,1) {$\cdots$};
\node (x11) at (-5,2) {$X^t_1$};
\node (x12) at (-4.5,1) {$X^t_2$};
\node (dot1) at (-4.2,0.2) {$\cdots$};
\node (x1n) at (-3.7,-0.6) {$X^t_n$};
\node (a1) at (-2,1) {$A^t$};
\node (x21) at (2,2) {$X^{t+1}_1$};
\node (x22) at (1.5,1) {$X^{t+1}_2$};
\node (dot2) at (1.2,0.2) {$\cdots$};
\node (x2n) at (0.7,-0.6) {$X^{t+1}_n$};
\node (a2) at (4,1) {$A^{t+1}$};
\node (u) at (-1,3) {$\U$};
\node (d2) at (6,1) {$\cdots$};

\path[->,thick] (x11) edge (a1);
\path[->,thick] (x1n) edge (a1);
\path[->,thick] (a1) edge (x21);
\path[->,thick] (a1) edge (x22);
\path[->,thick] (a1) edge (x2n);
\path[->,thick] (x21) edge (a2);
\path[->,thick] (x2n) edge (a2);
\path[->] (u) edge (x11);
\path[->] (u) edge (x12);
\path[->] (u) edge (x1n);
\path[->] (u) edge (x21);
\path[->] (u) edge (x22);
\path[->] (u) edge (x2n);
\path[dashed, ->] (x11) edge (x12);
\path[dashed, <-] (x12) edge (dot1);
\path[dashed, ->] (dot1) edge (x1n);
\path[dashed, ->] (x21) edge (x22);
\path[dashed, <-] (x22) edge (dot2);
\path[dashed, ->] (dot2) edge (x2n);
\end{tikzpicture}
}
\caption{An example causal graph of an imitation policy. At each time step $t$, the policy $\pi$ selects an action $A^t$ based on a subset of observations in $\X^t$. The resulting process consequently triggers new observations $\X^{t+1}$ together with the unmeasurable (hidden) exogenous variables $\U$. There may also exist (non-temporal) causal relationships within the observation space (an entangled representation), indicated by dashed arrows between elements in $\X$ (here, $X_2$ is partially caused by $X_1$). Causal relationships between the observation states and the policy are shown with solid arrows, so in this case $\X_2$ is part of the observation space but does not have a causal affect on the policy. The goal of this work is to learn these causal relationships from data.}
\label{fig:causal_graph}
\end{figure}
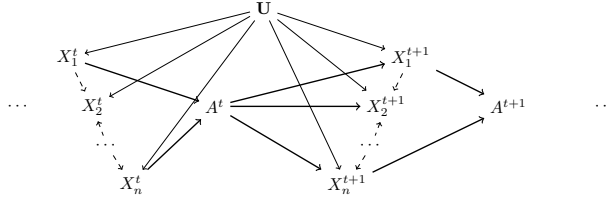


\paragraph{Causal graph}
We model the causal interactions between variables in imitation policies, i.e., observations and actions, in a graphic model.
Figure \ref{fig:causal_graph} shows an example of such causal relationships.
We use a {\em directed graph} to model the interactions between variables (i.e., observation dimensions and actions) involved in the policy, and we call such a directed graph a {\em causal graph} \cite{Pearl2000causal}.
In a causal graph, the nodes are random variables, e.g., observations and actions in Figure \ref{fig:causal_graph}, and a directed edge denotes a direct causal relationship between variables, e.g., $V_1\rightarrow V_2$ indicates that the change of variable $V_1$'s value directly causes the change of $V_2$'s value, but not vice versa.

In a causal graph $\g$, between two variables $V_1$ and $V_2$, if $V_1\rightarrow V_2$, $V_1$ is a {\em parent} of $V_2$, and we denote all parents of $V_2$ in as $pa_\g(V_2)$.
A sequence of nodes $(V_1, V_2, \cdots, V_m)$ is called a directed path if $V_i\rightarrow V_{i+1}$ for all $1\le i\le m-1$, and any other nodes on the directed path is called a {\em descendant} of $V_1$, and we denote $V_1$'s descendants as $de_\g(V_1)$, and the other nodes (i.e., non-descendants of $V_1$) as $nd_\g(V_1)$.
A causal graph induces the {\em local Markov property} of involved variables, i.e., for each variable $V$ in the causal graph:
\begin{equation}
V\perp nd_\g(V)\mid pa_\g(V).
\end{equation}
That is, $V$ is statistically independent from all its non-descendants conditioned on $V$'s parents.

Shaping an imitation policy $\pi$ as an above graphical causal model involving each time steps' observations and action, and unmeasurable exogenous variables as the nodes, we have a specific causal graph structure, e.g., one depicted in Figure \ref{fig:causal_graph}.
Concretely, for each time step $t$, $A^t$ has only in-degrees from a subset of $\X^t$, since $\pi$ decides $A^t$ only based on the corresponding observations.
That is, each observation dimension with an out-degree to the action has a direct influence on the decision of $A^t$.
Then, the next time step's observations $\X^{t+1}$ are consequently determined by the precursor action $A^t$ together with the exogenous variables $\U$.

Notice that the unmeasurable exogenous variable is not accessible to the agent, and thus there is no edge between $\U$ and $A^t$.
Importantly, in practice, it is not necessary that each dimension of the observation has influence on the agent's decision.
For instance, in the example introduced in Section \ref{sec:confusion}, the distracting cubes do not contain information of the task target, and thus the corresponding observation should not influence the decision.
Our objective in this work is to learn which dimensions of the observation are direct causes of the agent's decision, to prevent the agent from making biased decision by taking uninformed information.

Notice also that in Figure \ref{fig:causal_graph}, we allow the existence of causal relationships between dimensions of the observation at a time step\footnote{Note that such causal relationships between observation dimensions are not temporal as the relationship between $\X^t$ and $A^t$. Such non-temporal causal relationship can be revealed via observational data \cite{Pearl2000causal}.}, denoted as dashed edges.
This is different from one of the main claims in \cite{de2019causal}, which requires disentangled representation of the observation in the learning of the causal model.
{\bf In practice, learning causally disentangled representation is challenging \cite{komanduri2024learning}, and it is almost impossible without introducing biases \cite{locatello2019challenging}.}
Therefore, we theoretically address this problem in Section \ref{sec:condition}, by showing that we are guaranteed to learn a consistent structural function (defined as follows) without the disentangled representation condition.

\paragraph{Structural causal model}
Considering the causal graph modeling the imitation policy $\pi$, e.g., the one in Figure \ref{fig:causal_graph}, we assume that at each time step, the observation dimensions are fixed, e.g., each dimension denotes the information of the same graphical or semantic component of the observation.
Therefore, we restrict our analysis to a single time step, that is, learning the causal relationship between variables of $\X^t\cup\{A^t\}$.

Based on a causal graph, we further assume that a {\em structural causal model} (SCM) exists.
That is, the variables in the causal graph subject to counterfactuals with independent errors, i.e., the variables satisfy a data generation process:
for each variable $V$ in the causal graph, its value is decided by a {\em structural function}:
\begin{equation}
A = f_V(pa_\g(V), \epsilon_V),
\end{equation}
where the noise term $\epsilon_V$ is usually introduced due to the unmeasurable exogenous variables $\U$, and we assume the noise terms are mutually independent for each variable pair.
Then, an SCM is a tuple $\mathcal{M}=\langle \V, \U, \mathbf{\mathcal{V}}, \mathbf{\mathcal{U}}, \f, \mathbb{P}_{\mathbf{\mathcal{U}}}\rangle$, where $\V=\X^t\cup\{A^t\}$ is the set of endogenous variables (i.e., observable/measurable variables), $\U$ is the set of (unmeasurable/hidden) exogenous variables, $\mathbf{\mathcal{V}}$ is the product of domains of $\{V\}_{V\in \V}$, $\mathbf{\mathcal{U}}$ is the product of domains of $\{U\}_{U\in \U}$, $\f$ is the set of structural functions, and $\mathbb{P}_{\mathbf{\mathcal{U}}}$ is the product of exogenous distributions.
Intuitively, the structural function determines how observational data of the variables is generated.
For example, the demonstration sequences are generated based on expert's recognition of the cause structure, and thus this data is said to be {\em faithful} to the SCM.

Then, formally, in this work, we aim at first specifying the subset of observation dimensions in $\X^t$ that directly causes the decision of $A^t$ (i.e., $pa_\g(A^t)$), and then learning the mapping from $pa_\g(A^t)$ to the agent action $A^t$:
\begin{equation}
\label{eq:objective}
A^t = \hat{f}_\theta (pa_\g(A^t), \epsilon_{A^t}),
\end{equation}
that consistently estimates the true structural function $A^t = f_{A^t}(pa_\g(A^t), \epsilon_{A^t})$, with a series of expert demonstration observations and actions.

\subsection{Learnability of structural functions without disentangled representation requirement}
\label{sec:condition}
\cite{de2019causal} also tries to enhance imitation learning algorithms' generalization ability by learning a similar causal structure, however, they claim that a necessary condition for the framework is a disentangled representation of the observation, i.e. a mutually independent relationship between observation dimensions.
Apparently, forcing the mutual independence condition yields a simple causal graph containing no cycles, which generates a uniquely distributed dataset.
Then, an estimation based on such a dataset in turn identically approximates the underlying structural function.
However, when the causal relationships are complex, especially when cycles exist in the causal graph, a single SCM may generate multiple distributions, making it impossible to identify the underlying structural function.
However, in the case of imitation learning, due to the specific feature of the causal graph as specified in Section \ref{sec:causal_model}, we theoretically justify that causal connections, even cycles, among observation dimensions does not obstruct identifying the underlying structural function $f_{A^t}$.

Formally, we would like the imitation policy causal graph to satisfy certain conditions, such that if we properly estimate a mapping $g_{A^t}: pa_\g(A^t)\rightarrow A^t$, then this mapping uniquely indicates the underlying structural function, i.e., $$g_{A^t} \Longleftrightarrow f_{A^t}.$$
As follows, we formally introduce the function learnability property, and prove that it is satisfied in imitation policy causal graphs even if we discard the disentangled representation condition.

\paragraph{Unique Solvability}
We use the concept of (unique) solvability \cite{bongers2021foundations} to demonstrate that when we learn a mapping between a subset of variables, whether this mapping coincidentally interacts with the full SCM.
Note that we use the more restricted definition, i.e., unique solvability, as follows.
\begin{definition}[Unique Solvability \cite{bongers2021foundations}]
An SCM $\mathcal{M}=\langle \V, \U, \f, \mathbb{P}_{\mathbf{\xi}}\rangle$ is uniquely solvable w.r.t. $\V'\subseteq \V$ if there exists a measurable mapping $g_{\V'}: \mathbf{\mathcal{V}}_{pa_\g(\V')\setminus \V'}\times \mathbf{\mathcal{U}}_{pa_\g(\V')}\rightarrow \mathbf{\mathcal{V}}_{pa_\g(\V')}$ such that for $\mathbb{P}_{\mathbf{\mathcal{U}}}$-almost every $\mathbf{\epsilon}\in \mathbf{\mathcal{U}}$ and for all $\mathbf{v}\in \mathbf{\mathcal{V}}$,
$$\mathbf{v}_{\V'} = g_{\V'}(\mathbf{v}_{pa_\g(\V')\setminus \V'}, \mathbf{\epsilon}_{pa_\g(\V')}) \Longleftrightarrow \mathbf{v}_{\V'} = f_{\V'}(\mathbf{v}, \mathbf{\epsilon}).$$
\end{definition}

We refer to the following well recognized definition of disentangled representation to formalize the disentangled representation requirement for subsequent analysis.
\begin{definition}[Disentangled representation \cite{bengio2013representation}]
Disentangled representation should separate the distinct, independent and informative generative factors of variation in the data. Single latent variables are sensitive to changes in single underlying generative factors, while being relatively invariant to changes in other factors.
\end{definition}
Based on this definition, we formalize the requirement as: at each time step $t$, for each pair of observation dimensions $X^t_i, X^t_j\in \X^t$, they are mutually independent conditioned on the action at $t-1$, i.e.,
$$X^t_i \perp X^t_j\mid A^{t-1}.$$
However, as follows, we prove that this conditional independence condition is not a necessary requirement for unique solvability.


\begin{proposition}\label{prop:solvability}
A SCM $\mathcal{M}=\langle \V, \U, \f, \mathbb{P}_{\mathbf{\xi}}\rangle$ corresponding to a imitation policy causal graph is uniquely solvable w.r.t. $\{A^t\}$.
\end{proposition}



With relaxing of the disentangled representation requirement, theoretically, we may simply apply the causal structure learning component to a wide range of complex imitation learning algorithms for generalization enhancement.
In the next section, we show this simple strategy of incorporating the causal structure learning component in a recent imitation learning algorithm called Action Chunking Transformer (ACT) \cite{zhao2023learningACT}, to the strategy's efficacy.

\begin{figure}[tb]
  \centering
  \includegraphics[width=0.8\textwidth]{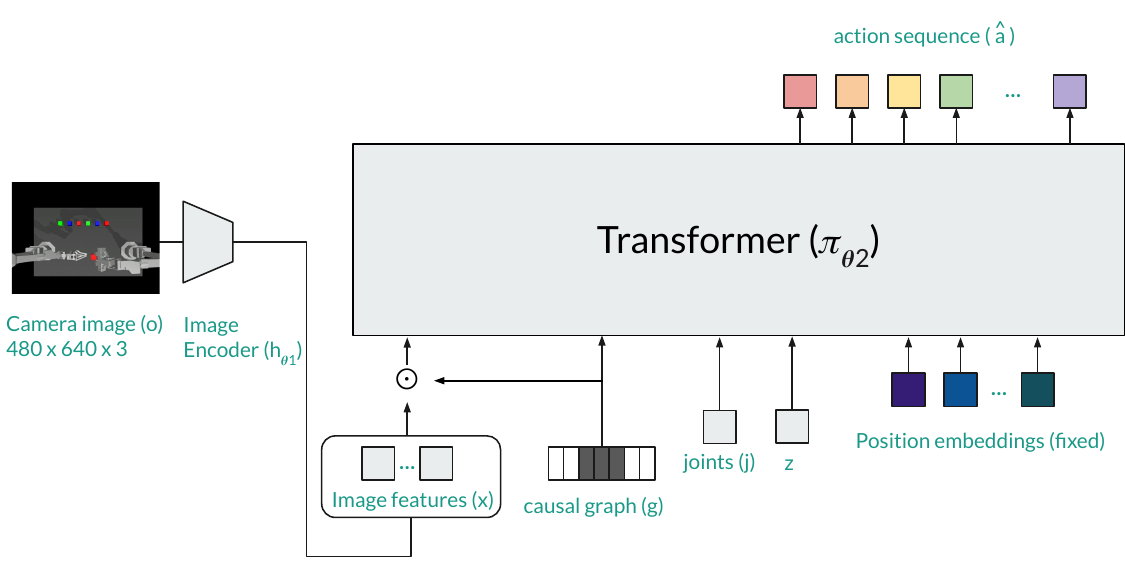}
  \vspace{-14pt}
  \caption{Model Architecture of \underline{Causal} Structure Learned \underline{ACT} (Causal-ACT). During training, a causal graph is uniformly sampled to modulate the image features. At test time, the causal graph is fixed to the best-performing graph which is decided by the targeted intervention process. As in ACT, the style variable $z$ is learned by a transformer encoder and set to zero at test time.}
  \label{fig:causal_act}
\end{figure}
\section{\texorpdfstring{Causal-ACT: \underline{Causal} Structure Learned \underline{ACT}}{Causal-ACT: Causal Structure Learned ACT}}
\label{sec:method}
This section introduces our proposed \underline{Causal} Structure Learned \underline{ACT} (Causal-ACT). To solve the causal confusion by finding the true causal structure, we intervene the image embedding input of the trained policy to collect interventional data.
This strategy is introduced in~\cite{de2019causal}.
First, we train a causal graph-parameterized ACT where various randomly sampled causal graphs are learned jointly with the ACT. Then, we perform the targeted intervention to find the causal graph that is expected to optimize performance.

In our work, Causal-ACT is built based on the ACT~\cite{zhao2023learningACT} by incorporating a causal graph as an additional input and using it to modulate the image features.
Figure~\ref{fig:causal_act} shows the model architecture of the Causal-ACT. The workflow of Causal-ACT is described in Algorithm~\ref{alg:causalACT}. 
Specifically, we use a ResNet encoder $h_{\theta_1}$ to extract image features $x_t$ based on the input image observation $o_t$ at timestep $t$. Then, the image features are multiplied element-wise with a candidate causal graph $g_t$, which is randomly sampled during training.
In particular, the causal graph is represented as an array of binary variables, and the array's dimensionality matches that of the image features. During training, each element of the graph is sampled uniformly i.i.d. As in ACT~\citep{zhao2023learningACT}, a style variable $z$ is learned by a transformer encoder $q_{\phi}(z|a_{t:t+k}, j_t)$, where $k$ is the chunk size of the action sequence and $j_t$ represents joint positions of the robot's arms.
The action sequence at time $t$ is predicted based on the graph-parameterized image features ($x_t*g_t$), the candidate graph ($g_t$), robot's joints ($j_t$), and the style variable ($z$), that is, $\hat{a}_{t:t+k}$ from $\pi_{\theta_2}(\hat{a}_{t:t+k} | x_t*g_t, g_t, j_t, z)$. In each training step, we perform a gradient descent step to minimize the loss function as shown in Equation~\ref{eq:loss}.

\begin{equation}\label{eq:loss}
    \mathcal{L} = MSE(\hat{a}_{t:t+k}, a_{t:t+k}) + \beta D_{KL}(q_{\phi}(z|a_{t:t+k}, j_t)) || \mathcal{N}(0, I)
\end{equation}

\noindent where $a$ and $\hat{a}$ represent the demonstrated action and the predicted action, respectively.
Note that parameters $\theta_1$ and $\theta_2$ are updated concurrently based on the MSE loss.

After training, we perform the targeted intervention according to Algorithm~\ref{alg:causalACT} to search for the graph that performs the best (denoted as $g^*$) according to the episodic reward. Here, one episode refers to the entire trajectory of performing the task from the initial to the final timestep. 
In our experiment, we follow the policy execution intervention method as described in~\citep{de2019causal}, where they use a linear energy-based model to learn the proper causal graph $g^*$.
Intuitively, the probability that a graph is the true graph is proportional to the episodic reward returned by executing the policy based on the graph.

When inferencing Causal-ACT, the causal graph $g$ is set to the best graph $g^*$, and the style variable $z$ is set to $0$. 
In other words, at inference timestep $t$, the predicted action sequence of size $k$ is $\hat{a}_{t:t+k}$ = $\pi_{\theta_2}(\hat{a}_{t:t+k} | x_t*g^*, g^*, j_t, 0)$, where $x_t = h_{\theta_1}(x_t \mid o_t)$.

\begin{algorithm}[t]
\caption{Causal Structure Learned Action Chunking Transformers (Causal-ACT): }
\label{alg:causalACT}
\begin{algorithmic}[1]
\STATE{Input expert demonstration dataset $\mathcal{D}$, chunk size $k$, weight $\beta$, \\and number of training epochs $n\_epochs$}
\STATE{Let $a_t$, $j_t$, $o_t$, and $g_t$ denote the action, robot proprioception, image observation, and sampled causal graph, respectively, at timestep $t$}
\STATE{Initialize image encoder $h_{\theta_1}(x_t\mid o_t)$, style encoder $q_{\phi}(z|a_{t:t+k}, j_t)$ and policy $\pi_{\theta_2}(\hat{a}_{t:t+k}\mid x_t*g_t, g_t, j_t, z)$}
\STATE{Initialize $\omega=0$, $D=\emptyset$}
    \FOR{iteration $n=1, 2, ..., n\_epochs$}
        \STATE Sample $o_t$, $j_t$, $a_{t:t+k}$ from $\mathcal{D}$\\
        \STATE Sample $z$ from $q_{\phi}(z|a_{t:t+k}, j_t)$\\
        \STATE Sample $g_t$ from $\mathcal{U}^{\mathrm{dim}(x)}\left\{ 0, 1 \right\}$\\
        \STATE Predict $x_t$ from $h_{\theta_1}(x_t\mid o_t)$
        \STATE Predict $\hat{a}_{t:t+k}$ from $\pi_{\theta_2}(\hat{a}_{t:t+k} | x_t*g_t, g_t, j_t, z)$
        \STATE $\mathcal{L}_{reconst} = MSE(\hat{a}_{t:t+k}, a_{t:t+k})$ \\
        \STATE $\mathcal{L}_{reg} = D_{KL}(q_{\phi}(z|a_{t:t+k}, j_t)) || \mathcal{N}(0, I)$
        \STATE Perform a gradient descent step on $\mathcal{L}=\mathcal{L}_{reconst} + \beta\mathcal{L}_{reg}$ \\with respect to the network parameters $\theta_1$, $\theta_2$, and $\phi$
    \ENDFOR

    \FOR{$i=1, ..., N$}
        \STATE Sample $g \sim p(g) \propto exp<\omega,g>$\\
        \STATE Compute reward $R_g$ by executing $\pi_{\theta_2}(\hat{a}_{t:t+k} | x_t*g, g, j_t, z)$ with trained image encoder $h_{\theta_1}(x_t\mid o_t)$ and style variable $z=0$\\
        \STATE $D \leftarrow D \cup \{(g, R_g)\}$\\
        \STATE Fit $\omega$ on $D$ with linear regression
    \ENDFOR
    \RETURN $ g^* = \arg \max_{g} p(g)$, $h_{\theta_1}$, and $\pi_{\theta_2}$
\end{algorithmic}
\end{algorithm}

\section{Experiments}\label{sec:experiment}
To empirically evaluate the effectiveness of our proposed method, Causal-ACT, we conduct simulated experiments on a set of bimanual robotic manipulation tasks using the ALOHA platform~\citep{zhao2023learningACT} and MuJoCo simulator~\citep{todorov2012mujoco}. Our experiments were chosen to test generalization under environmental domain shifts, specifically focusing on the impact of causal confusion on performance. We compare our method against two baselines, the original ACT algorithm and domain randomization strategies.

\subsection{Experimental Setup}\label{subsec:expsetups}
\paragraph{Simulated hardware and software}
ALOHA consists of two 6-DoF ViperX arms, configured for bimanual manipulation. All experiments are conducted in MuJoCo version 2.3.7 with top-view RGB camera inputs of resolution 480\texttimes640 with a control rate of 50 Hz. We simulate manipulation tasks on a laptop running Ubuntu 22.04 with Python 3.8, 64 GB RAM, a 13th Gen Intel Core i7-13850HX CPU, and an NVIDIA RTX 2000 Ada GPU.

\paragraph{Task Description}
We use a variant of the \textit{Cube Transfer} task~\citep{zhao2023learningACT}. In this task the right arm must pick up a red cube and transfer it to the left arm.
The cube's initial location is uniformly randomly sampled within an area of $20\text{cm}$\texttimes$20\text{cm}$, as shown by the red square in Figure~\ref{fig:env}. The task allows partial completion and an associated reward from 1 to 4 based on
whether the arm can \emph{touch} the cube, \emph{lift} it, \emph{attempt to transfer}, and \emph{complete the transfer} from one end effector to the other.

For imitation learning  purposes we gather a series of demonstrative `expert' trajectories using scripted end-effector control with full knowledge of the ground-truth state. Each of these training episodes consists of $400$ steps ($8$ s) and is stored as an HDF5 file ($\sim370$ MB each). This dataset comprises trajectories that include both joint and image data and is used to train the imitation algorithms.


\paragraph{Training vs Test Distribution}
To introduce causal confusion, six irrelevant distracting cubes with fixed color and position are added during training as shown in the left panel of Figure~\ref{fig:env}. 
Then, to examine the model's robustness, the trained agent is evaluated in an out-of-distribution (OOD) environment where there are no distracting cubes, as shown in the right panel of Figure~\ref{fig:env}. This simulates a domain shift where the environmental variables may appear to change, but the task remains the same.

\paragraph{Baselines}
In this work, we compare against two baseline methods, the original ACT algorithm and ACT trained with domain randomization. For the domain randomization method, ACT is trained with demonstrations containing a varying number, position and color of distractors. This setup follows conventional DR setups. We sample a variety of training environments with 1-6 distractors using a power weighted discrete distribution $P(i)=\frac{i^k}{\sum_{j=1}^k j^k},$. Where $k$ controls the skew of the distribution. Larger values of $k$ result in examples with higher numbers of cubes, which are further from the test distribution. Similarly, $k=0$ yields a uniform sampling of the number of cubes, which is closer to the test environment  (see Figure~\ref{fig:cube_sampling_dist} in Appendix~\ref{app:cube_sampling_dist}). Because the test environment contains no distractor cubes, environments with fewer cubes are closer to the test distribution.

\paragraph{Training Details}
Each ACT model has approximately 84M parameters. Training 2000 epochs takes around 55 minutes using 4GB of VRAM.
All models are trained on the same expert dataset (unless otherwise stated) with three random seeds. Inference of the models is real-time (50 Hz generating a chunk of 100 steps).
The Causal-ACT model is trained for 2000 epochs and takes around 65 minutes using 5GB of VRAM.
The hyperparameters of training ACT and Causal-ACT are shown in Appendix~\ref{app:param}.

\section{Results}
\paragraph{Generalization Performance}
Table~\ref{tab:act_eval} presents results of ACT and our Causal-ACT methods trained on the distractor training environment. Here we discuss the complete task (`transfer`) results unless noted otherwise. While ACT performs well in the in-distribution environment (0.89), its performance degrades substantially when tested out-of-distribution (0.23). This experiment shows an example of causal confusion: the ACT model has learned spurious correlations between the distractors and the task performance. Contrastingly, our Causal-ACT method maintains strong performance in-distribution (0.96) and substantially improves out-of-distribution performance (0.88) up from (0.23). This improved performance is achieved without needing additional expert training data or modeling additional environment variance. It was learned by learning to ignore the irrelevant components of the observation that do not lead to task performance. 

\begin{table}[t]
\caption{Success rates for ACT and Causal-ACT across distribution shifts. Each result is averaged over three seeds and 50 evaluation episodes.}
\label{tab:act_eval}
\centering
    \begin{tabular}{lcccccc}
        \toprule
        \textbf{Method} & \multicolumn{3}{c}{\textbf{Out-of-Distribution}} & \multicolumn{3}{c}{\textbf{In-Distribution}} \\
        \cmidrule(){2-7}
        & Touched & Lifted & Transfer & Touched & Lifted & Transfer \\
        \midrule
        ACT         & 0.87 & 0.54 & 0.23 & 0.99 & 0.96 & 0.89 \\
        Causal-ACT  & 0.88 & 0.84 & 0.82 & 0.96 & 0.96 & 0.96 \\
        \bottomrule
    \end{tabular}
\end{table}

\begin{table}[t]
    \caption{Success rates of plain ACT (row 1), Domain randomization with different sampling distributions (row 2-5), Causal-ACT (row 6), and ablation study of Causal-ACT with random causal graphs (row 7-8), for Out-of-Distribution tests. Each success rate is averaged over performances of three trained ACT agents that are trained with three different random seeds. Each agent is trained for $2000$ epochs and evaluated for $50$ episodes.}
    \label{tab:randomization}
    \centering
    \begin{tabular}{llll}
        \midrule
        \multicolumn{1}{c}{\multirow{2}{*}{Method}} & \multicolumn{3}{c}{Success rate}    \\ \cmidrule(){2-4} 
        \multicolumn{1}{c}{}  & \multicolumn{1}{c}{Touched} & \multicolumn{1}{c}{Lifted} & \multicolumn{1}{c}{Transfer} \\ \cmidrule(){1-4} 
        ACT                     & 0.87   & 0.54   & 0.23  \\
        ACT + DR ($k=+\infty$)  & 0.75   & 0.59   & 0.34         \\
        ACT + DR ($k=6$)        & 0.87   & 0.7    & 0.48       \\
        ACT + DR ($k=3$)        & 0.98   & 0.91   & 0.61       \\
        ACT + DR ($k=0$)        & 1.0    & 0.98   & 0.91     \\
        Causal-ACT              & 0.88   & 0.84   & 0.82    \\
        Causal-ACT (random graph)       & 0.91   & 0.82   & 0.48    \\
        Causal-ACT (full-connection graph)     & 0.1    & 0.08   & 0.02    \\
        \bottomrule
    \end{tabular}
\end{table}
\paragraph{Comparison to domain randomization}
Domain randomization can improve the performance if the sampled training environments overlap with the test environment enough. Our results for various $k$ values are summarized in Table~\ref{tab:randomization}.

Domain randomization has a varying level of performance based on how close the randomly sampled training domains are to the test domain. As expected, lower $k$ values (higher likelihood of fewer distractor cubes) correspond to increased performance (0.91) because the sampled environments are more similar to the test domain. Conversely, the domains with the largest train to test variance $k=\infty$ have similar performance to the models not using domain randomization. While domain randomization can improve performance, it requires extensive sampling and the assumption that the domain configurations and the sampled domains during training are relevant to testing conditions. Furthermore, randomization over irrelevant features can degrade performance.

Causal-ACT achieves a competitive level of performance with the best DR models while using significantly less data, requiring no manual domain design, and avoiding the issue of over-randomization.
Our results suggest that causal structure learning is a data-efficient and robust solution to generalization in imitation learning. We believe this is particularly valuable for real-world conditions where collecting additional data with domain randomization can be prohibitively expensive.

\paragraph{Ablation study}
We conduct an ablation study by sampling a causal graph at random (row 7, Table \ref{tab:act_eval}), and by using a fully connected graph between observation dimensions and action (row 8, Table \ref{tab:act_eval}).
Both experiments generate significantly lower performances (0.48 and 0.02) compared to Causal-ACT (row 7, Table \ref{tab:randomization}), showing the efficiency of Causal-ACT's graph searching process.

It is worth noticing that the success rate of the full-connection graph is considerably low (0.02).
We conjecture that Causal-ACT suffers from poor performance when the complete observation space is passed, both due to the high amount of irrelevant data and the lack of training in this regime.

\section{Conclusion}
\label{sec:con}
In this work, we addressed the problem of causal confusion in Imitation Learning (IL), where models rely on spurious correlations that impact their ability to generalize to new environments. Unlike prior work that depended on disentangled representations, we show that causal structure learning can be integrated directly into IL models. We developed Causal-ACT, a transformer-based IL framework that incorporates causal graph optimization to allow the policy to focus on task-relevant features. Our approach improves generalisation and sample efficiency, outperforming the ACT baseline and domain randomisation. 
Future work will extend this method to other IL architectures and real-world tasks, and explore alternative causal learning techniques to further enhance generalization.

\paragraph{Limitations} We would like to specify two limitations of the work.
Due to the general high dimensions of image inputs or their embedded representations, e.g. in our case, the ResNet encodes images into $n=512*15*20$ dimensional arrays, it is intractable to visit each possible graph from the large space (with size $2^n$).
Then, an efficient graph sampling method is desirable to differentiate good options fast, and this is one of our future works.



\bibliographystyle{plainnat}
\bibliography{robot_causal}  

\begin{thebibliography}{31}
\providecommand{\natexlab}[1]{#1}
\providecommand{\url}[1]{\texttt{#1}}
\expandafter\ifx\csname urlstyle\endcsname\relax
  \providecommand{\doi}[1]{doi: #1}\else
  \providecommand{\doi}{doi: \begingroup \urlstyle{rm}\Url}\fi

\bibitem[Anderson et~al.(2021)Anderson, Shrivastava, Truong, Majumdar, Parikh, Batra, and Lee]{anderson2021sim}
Peter Anderson, Ayush Shrivastava, Joanne Truong, Arjun Majumdar, Devi Parikh, Dhruv Batra, and Stefan Lee.
\newblock Sim-to-real transfer for vision-and-language navigation.
\newblock In \emph{Conference on Robot Learning}, pages 671--681. PMLR, 2021.

\bibitem[Bengio et~al.(2013)Bengio, Courville, and Vincent]{bengio2013representation}
Yoshua Bengio, Aaron Courville, and Pascal Vincent.
\newblock Representation learning: A review and new perspectives.
\newblock \emph{IEEE transactions on pattern analysis and machine intelligence}, 35\penalty0 (8):\penalty0 1798--1828, 2013.

\bibitem[Black et~al.(2024)Black, Brown, Driess, Esmail, Equi, Finn, Fusai, Groom, Hausman, Ichter, et~al.]{black2410pi0}
Kevin Black, Noah Brown, Danny Driess, Adnan Esmail, Michael Equi, Chelsea Finn, Niccolo Fusai, Lachy Groom, Karol Hausman, Brian Ichter, et~al.
\newblock $\pi$0: A vision-language-action flow model for general robot control.
\newblock \emph{URL https://arxiv. org/abs/2410.24164}, 2024.

\bibitem[Bongers et~al.(2021)Bongers, Forr{\'e}, Peters, and Mooij]{bongers2021foundations}
Stephan Bongers, Patrick Forr{\'e}, Jonas Peters, and Joris~M Mooij.
\newblock Foundations of structural causal models with cycles and latent variables.
\newblock \emph{The Annals of Statistics}, 49\penalty0 (5):\penalty0 2885--2915, 2021.

\bibitem[Brohan et~al.(2023)Brohan, Brown, Carbajal, Chebotar, Chen, Choromanski, Ding, Driess, Dubey, Finn, Florence, Fu, Arenas, Gopalakrishnan, Han, Hausman, Herzog, Hsu, Ichter, Irpan, Joshi, Julian, Kalashnikov, Kuang, Leal, Lee, Lee, Levine, Lu, Michalewski, Mordatch, Pertsch, Rao, Reymann, Ryoo, Salazar, Sanketi, Sermanet, Singh, Singh, Soricut, Tran, Vanhoucke, Vuong, Wahid, Welker, Wohlhart, Wu, Xia, Xiao, Xu, Xu, Yu, and Zitkovich]{brohanRT2VisionLanguageAction2023}
Anthony Brohan, Noah Brown, Justice Carbajal, Yevgen Chebotar, Xi~Chen, Krzysztof Choromanski, Tianli Ding, Danny Driess, Avinava Dubey, Chelsea Finn, Pete Florence, Chuyuan Fu, Montse~Gonzalez Arenas, Keerthana Gopalakrishnan, Kehang Han, Karol Hausman, Alexander Herzog, Jasmine Hsu, Brian Ichter, Alex Irpan, Nikhil Joshi, Ryan Julian, Dmitry Kalashnikov, Yuheng Kuang, Isabel Leal, Lisa Lee, Tsang-Wei~Edward Lee, Sergey Levine, Yao Lu, Henryk Michalewski, Igor Mordatch, Karl Pertsch, Kanishka Rao, Krista Reymann, Michael Ryoo, Grecia Salazar, Pannag Sanketi, Pierre Sermanet, Jaspiar Singh, Anikait Singh, Radu Soricut, Huong Tran, Vincent Vanhoucke, Quan Vuong, Ayzaan Wahid, Stefan Welker, Paul Wohlhart, Jialin Wu, Fei Xia, Ted Xiao, Peng Xu, Sichun Xu, Tianhe Yu, and Brianna Zitkovich.
\newblock {{RT-2}}: {{Vision-Language-Action Models Transfer Web Knowledge}} to {{Robotic Control}}, 2023.

\bibitem[De~Haan et~al.(2019)De~Haan, Jayaraman, and Levine]{de2019causal}
Pim De~Haan, Dinesh Jayaraman, and Sergey Levine.
\newblock Causal confusion in imitation learning.
\newblock \emph{Advances in neural information processing systems}, 32, 2019.

\bibitem[Ho and Ermon(2016)]{hoGenerativeAdversarial2016}
Jonathan Ho and Stefano Ermon.
\newblock Generative {{Adversarial Imitation Learning}}.
\newblock In \emph{Advances in {{Neural Information Processing Systems}}}, volume~29. Curran Associates, Inc., 2016.

\bibitem[Izmailov et~al.(2022)Izmailov, Kirichenko, Gruver, and Wilson]{izmailov2022feature}
Pavel Izmailov, Polina Kirichenko, Nate Gruver, and Andrew~G Wilson.
\newblock On feature learning in the presence of spurious correlations.
\newblock \emph{Advances in Neural Information Processing Systems}, 35:\penalty0 38516--38532, 2022.

\bibitem[James et~al.(2019)James, Wohlhart, Kalakrishnan, Kalashnikov, Irpan, Ibarz, Levine, Hadsell, and Bousmalis]{jamesSimToRealSimToSim2019}
Stephen James, Paul Wohlhart, Mrinal Kalakrishnan, Dmitry Kalashnikov, Alex Irpan, Julian Ibarz, Sergey Levine, Raia Hadsell, and Konstantinos Bousmalis.
\newblock Sim-{{To-Real}} via {{Sim-To-Sim}}: {{Data-Efficient Robotic Grasping}} via {{Randomized-To-Canonical Adaptation Networks}}.
\newblock In \emph{Proceedings of the {{IEEE}}/{{CVF Conference}} on {{Computer Vision}} and {{Pattern Recognition}}}, 2019.

\bibitem[Kingma and Ba(2014)]{Adam}
Diederik~P Kingma and Jimmy Ba.
\newblock Adam: A method for stochastic optimization.
\newblock \emph{arXiv preprint arXiv:1412.6980}, 2014.

\bibitem[Komanduri et~al.(2024)Komanduri, Wu, Chen, and Wu]{komanduri2024learning}
Aneesh Komanduri, Yongkai Wu, Feng Chen, and Xintao Wu.
\newblock Learning causally disentangled representations via the principle of independent causal mechanisms.
\newblock In \emph{Proceedings of the Thirty-Third International Joint Conference on Artificial Intelligence}, pages 4308--4316, 2024.

\bibitem[Locatello et~al.(2019)Locatello, Bauer, Lucic, Raetsch, Gelly, Sch{\"o}lkopf, and Bachem]{locatello2019challenging}
Francesco Locatello, Stefan Bauer, Mario Lucic, Gunnar Raetsch, Sylvain Gelly, Bernhard Sch{\"o}lkopf, and Olivier Bachem.
\newblock Challenging common assumptions in the unsupervised learning of disentangled representations.
\newblock In \emph{international conference on machine learning}, pages 4114--4124. PMLR, 2019.

\bibitem[Makoviychuk et~al.(2021)Makoviychuk, Wawrzyniak, Guo, Lu, Storey, Macklin, Hoeller, Rudin, Allshire, Handa, and State]{makoviychukIsaacGym2021}
Viktor Makoviychuk, Lukasz Wawrzyniak, Yunrong Guo, Michelle Lu, Kier Storey, Miles Macklin, David Hoeller, Nikita Rudin, Arthur Allshire, Ankur Handa, and Gavriel State.
\newblock Isaac {{Gym}}: {{High Performance GPU Based Physics Simulation For Robot Learning}}.
\newblock \emph{Proceedings of the Neural Information Processing Systems Track on Datasets and Benchmarks}, 1, 2021.

\bibitem[OpenAI et~al.(2019)OpenAI, Akkaya, Andrychowicz, Chociej, Litwin, McGrew, Petron, Paino, Plappert, Powell, Ribas, Schneider, Tezak, Tworek, Welinder, Weng, Yuan, Zaremba, and Zhang]{openaiSolvingRubiks2019}
OpenAI, Ilge Akkaya, Marcin Andrychowicz, Maciek Chociej, Mateusz Litwin, Bob McGrew, Arthur Petron, Alex Paino, Matthias Plappert, Glenn Powell, Raphael Ribas, Jonas Schneider, Nikolas Tezak, Jerry Tworek, Peter Welinder, Lilian Weng, Qiming Yuan, Wojciech Zaremba, and Lei Zhang.
\newblock Solving {{Rubik}}'s {{Cube}} with a {{Robot Hand}}, 2019.

\bibitem[Osa et~al.(2018)Osa, Pajarinen, Neumann, Bagnell, Abbeel, and Peters]{osaAlgorithmicPerspective2018}
Takayuki Osa, Joni Pajarinen, Gerhard Neumann, J.~Andrew Bagnell, Pieter Abbeel, and Jan Peters.
\newblock An {{Algorithmic Perspective}} on {{Imitation Learning}}.
\newblock \emph{Foundations and Trends{\textregistered} in Robotics}, 7\penalty0 (1-2), 2018.
\newblock ISSN 1935-8253, 1935-8261.
\newblock \doi{10.1561/2300000053}.

\bibitem[O’Neill et~al.(2024)O’Neill, Rehman, Maddukuri, Gupta, Padalkar, Lee, Pooley, Gupta, Mandlekar, Jain, et~al.]{o2024open}
Abby O’Neill, Abdul Rehman, Abhiram Maddukuri, Abhishek Gupta, Abhishek Padalkar, Abraham Lee, Acorn Pooley, Agrim Gupta, Ajay Mandlekar, Ajinkya Jain, et~al.
\newblock Open x-embodiment: Robotic learning datasets and rt-x models: Open x-embodiment collaboration 0.
\newblock In \emph{2024 IEEE International Conference on Robotics and Automation (ICRA)}, pages 6892--6903. IEEE, 2024.

\bibitem[Pearl(2000)]{Pearl2000causal}
Judea Pearl.
\newblock \emph{Causality: Models, Reasoning and Inference}.
\newblock Cambridge University Press, New York, 2000.

\bibitem[Peters et~al.(2017)Peters, Janzing, and Sch{\"o}lkopf]{peters2017elements}
Jonas Peters, Dominik Janzing, and Bernhard Sch{\"o}lkopf.
\newblock \emph{Elements of causal inference: foundations and learning algorithms}.
\newblock The MIT Press, 2017.

\bibitem[Prakash et~al.(2019)Prakash, Boochoon, Brophy, Acuna, Cameracci, State, Shapira, and Birchfield]{prakashStructuredDomain2019}
Aayush Prakash, Shaad Boochoon, Mark Brophy, David Acuna, Eric Cameracci, Gavriel State, Omer Shapira, and Stan Birchfield.
\newblock Structured {{Domain Randomization}}: {{Bridging}} the {{Reality Gap}} by {{Context-Aware Synthetic Data}}.
\newblock In \emph{2019 {{International Conference}} on {{Robotics}} and {{Automation}} ({{ICRA}})}, 2019.
\newblock \doi{10.1109/ICRA.2019.8794443}.

\bibitem[Shafiullah et~al.(2022)Shafiullah, Cui, Altanzaya, and Pinto]{shafiullahBehaviorTransformers2022}
Nur~Muhammad Shafiullah, Zichen Cui, Ariuntuya~(Arty) Altanzaya, and Lerrel Pinto.
\newblock Behavior {{Transformers}}: {{Cloning}} k modes with one stone.
\newblock \emph{Advances in Neural Information Processing Systems}, 35, 2022.

\bibitem[Shridhar et~al.(2023)Shridhar, Manuelli, and Fox]{shridharPerceiverActorMultiTask2023}
Mohit Shridhar, Lucas Manuelli, and Dieter Fox.
\newblock Perceiver-{{Actor}}: {{A Multi-Task Transformer}} for {{Robotic Manipulation}}.
\newblock In \emph{Proceedings of {{The}} 6th {{Conference}} on {{Robot Learning}}}. PMLR, 2023.

\bibitem[Targ et~al.(2016)Targ, Almeida, and Lyman]{targ2016resnet}
Sasha Targ, Diogo Almeida, and Kevin Lyman.
\newblock Resnet in resnet: Generalizing residual architectures.
\newblock \emph{arXiv preprint arXiv:1603.08029}, 2016.

\bibitem[Tobin et~al.(2017)Tobin, Fong, Ray, Schneider, Zaremba, and Abbeel]{tobinDomainRandomization2017}
Josh Tobin, Rachel Fong, Alex Ray, Jonas Schneider, Wojciech Zaremba, and Pieter Abbeel.
\newblock Domain randomization for transferring deep neural networks from simulation to the real world.
\newblock In \emph{2017 {{IEEE}}/{{RSJ International Conference}} on {{Intelligent Robots}} and {{Systems}} ({{IROS}})}, 2017.
\newblock \doi{10.1109/IROS.2017.8202133}.

\bibitem[Todorov et~al.(2012)Todorov, Erez, and Tassa]{todorov2012mujoco}
Emanuel Todorov, Tom Erez, and Yuval Tassa.
\newblock Mujoco: A physics engine for model-based control.
\newblock In \emph{2012 IEEE/RSJ International Conference on Intelligent Robots and Systems}, pages 5026--5033. IEEE, 2012.
\newblock \doi{10.1109/IROS.2012.6386109}.

\bibitem[Yao et~al.(2021)Yao, Chu, Li, Li, Gao, and Zhang]{yaoSurveyCausal2021}
Liuyi Yao, Zhixuan Chu, Sheng Li, Yaliang Li, Jing Gao, and Aidong Zhang.
\newblock A {{Survey}} on {{Causal Inference}}.
\newblock \emph{ACM Trans. Knowl. Discov. Data}, 15\penalty0 (5), 2021.
\newblock ISSN 1556-4681.
\newblock \doi{10.1145/3444944}.

\bibitem[Zare et~al.(2024)Zare, Kebria, Khosravi, and Nahavandi]{zareSurveyImitation2024}
Maryam Zare, Parham~M. Kebria, Abbas Khosravi, and Saeid Nahavandi.
\newblock A {{Survey}} of {{Imitation Learning}}: {{Algorithms}}, {{Recent Developments}}, and {{Challenges}}.
\newblock \emph{IEEE Transactions on Cybernetics}, 54\penalty0 (12), 2024.
\newblock ISSN 2168-2275.
\newblock \doi{10.1109/TCYB.2024.3395626}.

\bibitem[Zeng et~al.(2021)Zeng, Florence, Tompson, Welker, Chien, Attarian, Armstrong, Krasin, Duong, Sindhwani, and Lee]{zengTransporterNetworks2021}
Andy Zeng, Pete Florence, Jonathan Tompson, Stefan Welker, Jonathan Chien, Maria Attarian, Travis Armstrong, Ivan Krasin, Dan Duong, Vikas Sindhwani, and Johnny Lee.
\newblock Transporter {{Networks}}: {{Rearranging}} the {{Visual World}} for {{Robotic Manipulation}}.
\newblock In \emph{Proceedings of the 2020 {{Conference}} on {{Robot Learning}}}. PMLR, 2021.

\bibitem[Zhang et~al.(2020)Zhang, Lyle, Sodhani, Filos, Kwiatkowska, Pineau, Gal, and Precup]{zhang2020invariant}
Amy Zhang, Clare Lyle, Shagun Sodhani, Angelos Filos, Marta Kwiatkowska, Joelle Pineau, Yarin Gal, and Doina Precup.
\newblock Invariant causal prediction for block mdps.
\newblock In \emph{International Conference on Machine Learning}, pages 11214--11224. PMLR, 2020.

\bibitem[Zhao et~al.(2023)Zhao, Kumar, Levine, and Finn]{zhao2023learningACT}
Tony~Z Zhao, Vikash Kumar, Sergey Levine, and Chelsea Finn.
\newblock Learning fine-grained bimanual manipulation with low-cost hardware.
\newblock In \emph{Robotics: Science and Systems XIX}, Daegu, Republic of Korea, July 2023.
\newblock \doi{10.15607/RSS.2023.XIX.016}.

\bibitem[Zhao et~al.(2024)Zhao, Tompson, Driess, Florence, Ghasemipour, Finn, and Wahid]{zhao2024aloha}
Tony~Z Zhao, Jonathan Tompson, Danny Driess, Pete Florence, Kamyar Ghasemipour, Chelsea Finn, and Ayzaan Wahid.
\newblock Aloha unleashed: A simple recipe for robot dexterity.
\newblock \emph{arXiv preprint arXiv:2410.13126}, 2024.

\bibitem[Zhao et~al.(2020)Zhao, Queralta, and Westerlund]{zhao2020sim}
Wenshuai Zhao, Jorge~Pe{\~n}a Queralta, and Tomi Westerlund.
\newblock Sim-to-real transfer in deep reinforcement learning for robotics: a survey.
\newblock In \emph{2020 IEEE symposium series on computational intelligence (SSCI)}, pages 737--744. IEEE, 2020.

\end{thebibliography}

\newpage
\appendix

\section{Proof of Proposition \ref{prop:solvability}}

\begin{proof}
To prove Proposition \ref{prop:solvability}, we need the following lemma.
\begin{lemma}[\cite{bongers2021foundations}]\label{lem:solve_single}
A SCM $\mathcal{M}=\langle \V, \U, \f, \mathbb{P}_{\mathbf{\xi}}\rangle$ is uniquely solvable w.r.t. a single variable $\{V\}\subseteq \V$ if and only if $V$ has no self cycle in the corresponding causal graph, i.e., edge $V\rightarrow V$ does not exist.
\end{lemma}
According to the specification of a imitation learning causal graph, variable $A^t$ only has in-degrees from a subset of $\X^t$.
Then, the causal graph satisfies the condition required in Lemma \ref{lem:solve_single}, and therefore the SCM corresponding to the causal graph is uniquely solvable.
\end{proof}

\section{Hyperparameters of Training}
\label{app:param}

For a fair comparison, we keep the hyperparameters of training ACT and Causal-ACT the same as reported in ~\citep{zhao2023learningACT}. As is shown in Table~\ref{tab:hyperparam}, intervention iteration is the extra hyperparameter for Causal-ACT to get a best-performed graph. Both ACT and Causal-ACT are trained for $2000$ epochs with Adam \citep{Adam} of the standard values of $\beta_1 = 0.9$ and $\beta_2 = 0.999$ as the optimizer.

\begin{table}[h]
  \caption{Hyperparameters of training ACT and Causal-ACT.}
  \label{tab:hyperparam}
  \centering
  \begin{tabular}{rl}
    \toprule
    learning rate     & $10^{-5}$     \\
    chunk size & $100$ \\
    batch size     & $8$ \\
    number of heads & $8$ \\
    number of encoder layers & $4$ \\
    number of decoder layers & $7$ \\
    hidden dimension & $512$ \\
    feedforward dimension & $3200$ \\
    learning rate of backbone & $10^{-5}$ \\
    beta & $10$ \\
    intervention iteration & $50$ \\
    \bottomrule
  \end{tabular}
\end{table}

\section{Domain randomization sampling distributions}
\label{app:cube_sampling_dist}

\begin{figure}[h]
     \centering
     \includegraphics[width=0.24\textwidth]{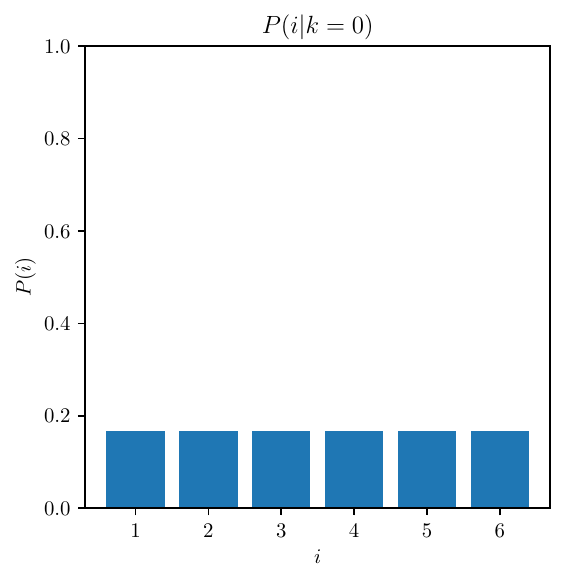}
      \includegraphics[width=0.24\textwidth]{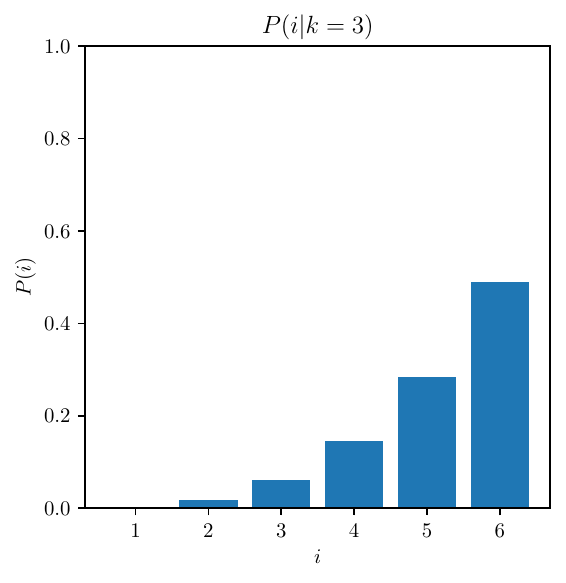}
       \includegraphics[width=0.24\textwidth]{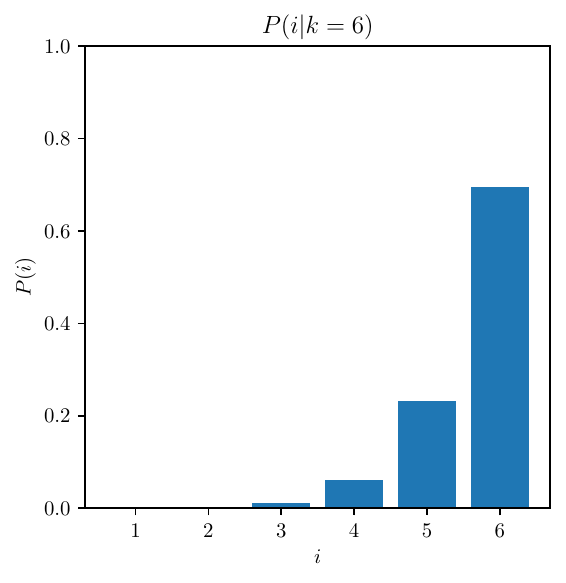}
    \includegraphics[width=0.24\textwidth]{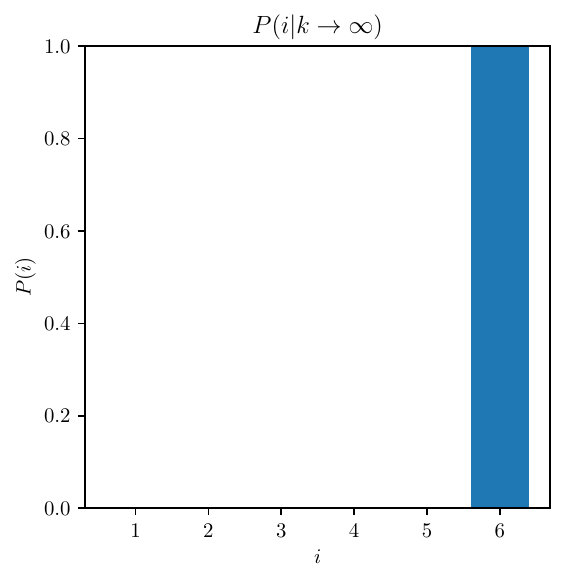}
     \caption{The distribution used to sample the number of cubes in domain randomization.}
     \label{fig:cube_sampling_dist}
\end{figure}

Figure~\ref{fig:cube_sampling_dist} illustrates the likelihood of sampling $i=1,\ldots,6$ cubes under the power weighted discrete distribution for values of $k \in \left\{0, 3, 6, \rightarrow \infty \right\}$.


\end{document}